\documentclass[twocolumn,10pt]{asme2ej}

\usepackage{graphicx} 
\usepackage{hyperref}   
\hypersetup{
	colorlinks=true,
	linkcolor=blue,
	citecolor=blue,
	urlcolor=blue,
}
\usepackage[square,numbers]{natbib}
\usepackage[caption=false]{subfig}
\usepackage[flushleft]{threeparttable}
\usepackage{import,threeparttable}
\usepackage{amsmath,amssymb}
\usepackage[latin1]{inputenc}
\usepackage[flushleft]{threeparttable}
\usepackage{import,threeparttable}
\usepackage{tabularx}
\usepackage{multirow}
\usepackage{algorithm,algorithmic}
\usepackage{xcolor}

\newcommand\crule[3][black]{\textcolor{#1}{\rule{#2}{#3}}}
\definecolor{darkgreen}{HTML}{006400}
\definecolor{darkred}{HTML}{CC0000}
\definecolor{yellow}{HTML}{FFCC00}
\definecolor{orange}{HTML}{FF6633}

\title{Unsupervised Movement Detection in Indoor Positioning Systems of Production Halls}

\author{Jonathan Flossdorf
    \affiliation{
	Department of Statistics\\
	TU Dortmund University\\
    Email: flossdorf@statistik.tu-dortmund.de
    }	
}

\author{Anne Meyer
    \affiliation{
	Faculty of Mechanical Engineering\\
	TU Dortmund University
    }	
}

\author{Dmitri Artjuch
    \affiliation{
	Department of Statistics\\
	TU Dortmund University
    }	
}

\author{Jaques Schneider
    \affiliation{
	Arnold AG, Friedrichsdorf
    }	
}

\author{Carsten Jentsch
    \affiliation{
	Department of Statistics\\
	TU Dortmund University
    }	
}

\begin{document}

\maketitle    

\begin{abstract}
{\it Consider indoor positioning systems (IPS) in production halls where objects equipped with sensors send their current position. Beside its large volume, the analyzation of the resulting raw data is challenging due to the susceptibility towards noise. Reasons are accuracy issues and undesired awakenings of sensors that occur due to the dynamics of logistic processes (e.g.~vibrations of passing forklifts). We propose a tailor-made statistical procedure for these challenges and combine visual analytics with movement detection. Contrary to common stay-point algorithms, we do not only distinguish between stops and moves, but also consider undesired awakenings. This leads to a more detailed interpretation scheme offering usages for online (e.g.~monitoring of orders) and offline applications (e.g.~detection of problematic areas). The approach does not require other information than the raw IPS output and enables an ad-hoc analysis. We underline our findings in an extensive case study with real IPS data of our industry partner.
}
\end{abstract}

\section{Introduction}

Indoor Positioning Systems (IPS) are a popular research field and suitable for various application fields \citep{mier2019glance} ranging from discovering customer behavior in shopping malls \citep{renaudin2019evaluating, pei2020influences} to improving indoor security at airports \citep{molina2018multimodal, zekavat2004novel}. There exist various survey papers of different technological approaches and potentials \citep{zafari2019survey, mainetti2014survey, al2011survey}. In this work, we focus on the particular application of IPS in production halls of manufacturing companies where satellites are attached at various points of the hall and communicate with mobile receivers equipped with sensors which can locate themselves. Thus, objects like vehicles, load carriers or components can be individually tracked. Beside obvious benefits like reduced search efforts, these systems have secondary advantages on assistance systems, fault management or production control \citep{mieth2019survey, mieth2019framework}. 

\subsection{Data and Problem Specification}
Depending on the used technology and settings of the IPS, the structure of the output data can be different in some details. However, the main characteristics are usually the same and involve a) the ID of the sensor, b) the position coordinates in three dimensions and c) a timestamp. If those characteristics are sent, we say that an \emph{event} took place. We denote the amount of sensors in the whole system as $K$ and use their ID as the index $j$. We denote the position vector of event $i$ for device $j$ as \text{\textbf{u$_{i,j}$}} $= (x_{i,j}, y_{i,j}, z_{i,j})'$ and the corresponding timestamp as $t_{i,j}$. The index $i = 1, ..., n_{j}$ signalizes the chronological order of the events for the corresponding device $j$. In this context, $n_{j}$ represents the number of events. We worked together with our industry partner Arnold AG that is a company of the sheet metal industry with an own IPS. Throughout the paper, we illustrate and evaluate our procedure with their real IPS data of over 3.5 million position events collected in a time period of two months. As we can see by those numbers, IPS are complex and their treatment is challenging. The sheer volume of resulting data is rather useless in its raw state, which is why suitable methods of data mining are crucial to make the data informative and actionable. The main focus typically lies on the understanding of the movement structure in the system, e.g. the detection of moves and stops. However, this task is made much more difficult due to the fact that IPS are usually quite prone to noise. In this context, noise is typically equated with accuracy issues of the data due to measuring errors \citep{zimmermann2009finding} that can amount to several meters \citep{al2011survey} despite of the development of different localization algorithms \citep{bahl2000radar, madigan2005bayesian, bekkali2007rfid, ali2009fingerprint}. This is particularly the case for production halls, since these are characterized by narrow regions and high-bay warehouses that both tend to have an increased negative impact on the signal quality. We note this noise type as ACC, but we underline that it is crucial to be aware of another type of noise that we introduce as SLEEP. In a nutshell, the type SLEEP describes events that were undesirably sent. Precisely, SLEEP is caused by the fact that most Internet of Things devices require long sleeping cycles in order to last for long periods \citep{zafari2019survey}. Those cycles are interrupted by wake-up periods whose occurrence may vary across different system specifications. In our research, the IPS of our industry partner is constructed such that sensors only send position events, when they detected a movement of the attached object. However, a frequent occurrence of false alarms, i.e. a sensor sends undesired position information, is likely. In practice, this is mainly caused by the dynamic nature of logistic processes through vibrations which activate the sensors, e.g. due to passing fork-lift trucks or working production machines nearby. While some work has been done regarding the consideration of ACC \citep{zimmermann2009finding, xiang2016extracting}, the handling of SLEEP was, to the best of our knowledge, not discussed in the literature so far. However, its consideration is crucial as undesirable events a) have a negative impact on the overall data quality causing a larger, more confusing and less informative IPS output, b) affect an increased energy consumption of the sensors which shortens their charging cycles and battery life leading to a decreased efficiency and c) may lead to wrong movement interpretations.  

\subsection{Contribution} \label{sec:Contr}
To this purpose, we develop a novel statistical procedure combining visual analytics with movement detection algorithms that is tailor-made for the described IPS challenges in production halls. We take both noise types into account and particularly aim to detect possible candidate events for undesired awakenings and distinguish them from actual movements and natural stops. Based on this, we design an interpretation scheme that hints to typical system characteristics and the actual movement structure. We illustrate its usage for various applications including:
\begin{itemize}
\item[(i)] Online Applications: Real-time monitoring of process flows in order to rapidly confirm that an order works as planned or to send an alarm if problems occur
\item[(ii)] Offline Applications: Development of a map that is automatically able to detect and to illustrate the current main characteristics including transport routes, warehouses or production machines that enables the user to e.g.:
\begin{itemize}
\item[-] detect problematic system areas with a high amount of undesired awakenings and to identify possible interference sources
\item[-] take suitable countermeasures for those areas like an adaption of sensibility settings of the sensors to the interference intensity of the corresponding local region in order to decrease the amount of undesired awakenings
\item[-] use the map as an update and validation  process for rather dynamic or season-dependent environments
\end{itemize} 
\end{itemize}
Due to the sheer volume of IPS data, the elevation of a ground truth is hardly feasible in practice. Thus, we only use unsupervised techniques in order to make the approach ad-hoc applicable and enable an analysis only based on the raw output data with no further information required.

\subsection{Related Work} \label{sec:relW}
Related IPS data mining methods involve an algorithm which finds the current most dense (i.e. busy) indoor areas from a set of user-defined query regions \citep{li2018search}. In a similar context, outlier detection was performed with a combination of supervised, unsupervised and ensemble machine learning methods in order to take the susceptibility of IPS towards measuring errors into account \citep{bhatti2020outlier}. Various works perform trajectory mining to analyze specific movement patterns of actors in the system. This e.g~involves the analyzation of similarities between observed trajectories with the support of density-based clustering approaches adapted to the challenges of indoor data \citep{cheng2021clustering}. The most related research question to our work is the detection of stops in a movement trajectory in order to detect important stay points. The algorithm SMoT \citep{alvares2007model} is designed to handle this task by dividing a trajectory into stops and moves, but requires a-priori information about the places of stops which is a rather strict assumption for IPS data. The adaption CB-SMoT \citep{palma2008clustering} solves this issue by applying density-based clustering algorithms and associating each detected cluster with a stop, but is having problems with stops after movements of varying speed that are typical for our considered application focus (e.g.~transportation with continuous or non-continuous conveyors). A further adaption considers frequent direction changes as an additional reference for a stop \citep{rocha2010db}. However, a better interpretable stay point algorithm for IPS data needs to consider the time component beside the pure distances as the observations naturally occur time-ordered. Otherwise two observations, that are lying close to each other, are assigned to the same stop cluster although they were part of two completely different movements. A solution is presented in \citep{zimmermann2009finding}, where an adaption of the density-based clustering algorithm OPTICS \citep{ankerst1999optics} to time-dependent data is proposed by the addition of a time proximity constraint to the already existing distance criterion that enables the user to analyze the stop behavior at different time points. Similarly, a sequence oriented clustering approach was developed which also captures the duration and not only proximity in the time component \citep{xiang2016extracting} and therefore allows for a more detailled analyzation. Note that the mentioned algorithms only aim to distinguish between stops and moves, whereas we focus on the further distinction between undesired sent data and actual movements. From our point of view, this is necessary for IPS in production halls as detected stops might actually be undesired awakenings in many cases. Contrary to the related literature, we therefore take noise type SLEEP specifically into account. From a technical view, we achieve this by analyzing the time and distance component separately and combine them afterwards with the help of visual analyzation to a more flexible interpretation scheme that enables the user to profit from the benefits described in Section \ref{sec:Contr}.\\
\\
The paper is organized as follows: In Section 2, we modify established movement algorithms to our purposes and explain their functioning in detail. Afterwards, we combine the information of the algorithms to a statistical procedure, design a comprehensive interpretation scheme and propose offline and online application scenarios. Our findings are underlined and illustrated by the use of an extensive real-world example in Section 3 that we examined and evaluated together with the IPS data and expertise of our industry partner Arnold AG. The final Section 4 consists of some concluding remarks. 

\section{Movement Classification} 
In order to derive a reliable and flexible interpretation scheme of the movement structure in production halls, we aim to distinguish between events, which were an actual movement of the attached object, and events that were undesirably sent (noise type SLEEP). We denote both classes as AME (Actual Movement Event) and UAE (Undesired Awakening Event), respectively. In order to distinguish between the AME and UAE class, we use all available information and two different unsupervised classification algorithm approaches. This involves a distance-based criterion, where small or no traveled distances after a certain number of measured steps indicate the UAE type, and a time-based criterion, where time-discrete awakenings (i.e. rather long intervals between single events of a device) are characteristic for the UAE type. Note that both algorithms use the well-established idea of density-based clustering and the distance-based algorithm is in its main idea similar to the ones in \citep{zimmermann2009finding, xiang2016extracting} and is therefore considering flexible intensities of noise type ACC as well. However, as explained above, we have a different application focus, require some modifications for the consideration of noise type SLEEP and use a separate analyzation of the distance and time-based algorithm to combine them to a more comprehensive classification allowing for a more flexible movement interpretation scheme.
  
\begin{figure*}[!t]
\centering
\subfloat[Actual Movement]{\includegraphics[width=2.7in]{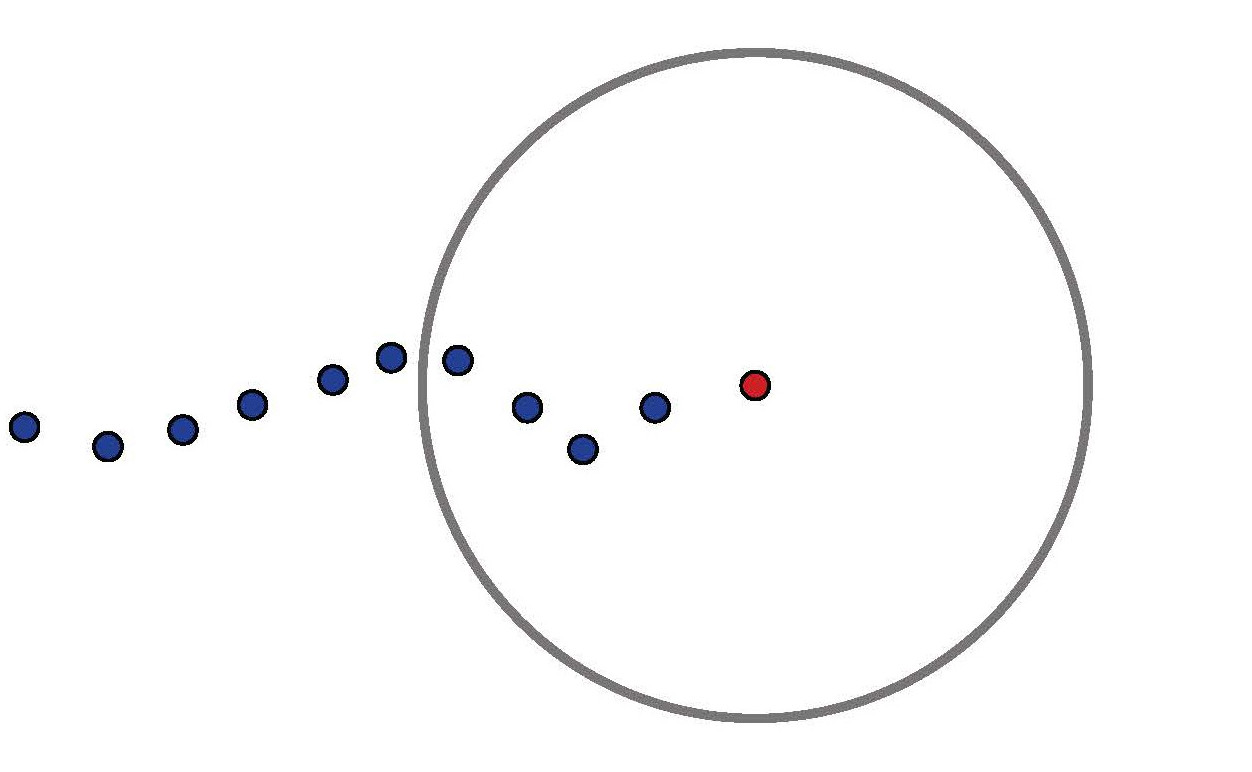}}%
\hfil
\subfloat[Undesired Awakening]{\includegraphics[width=2.7in]{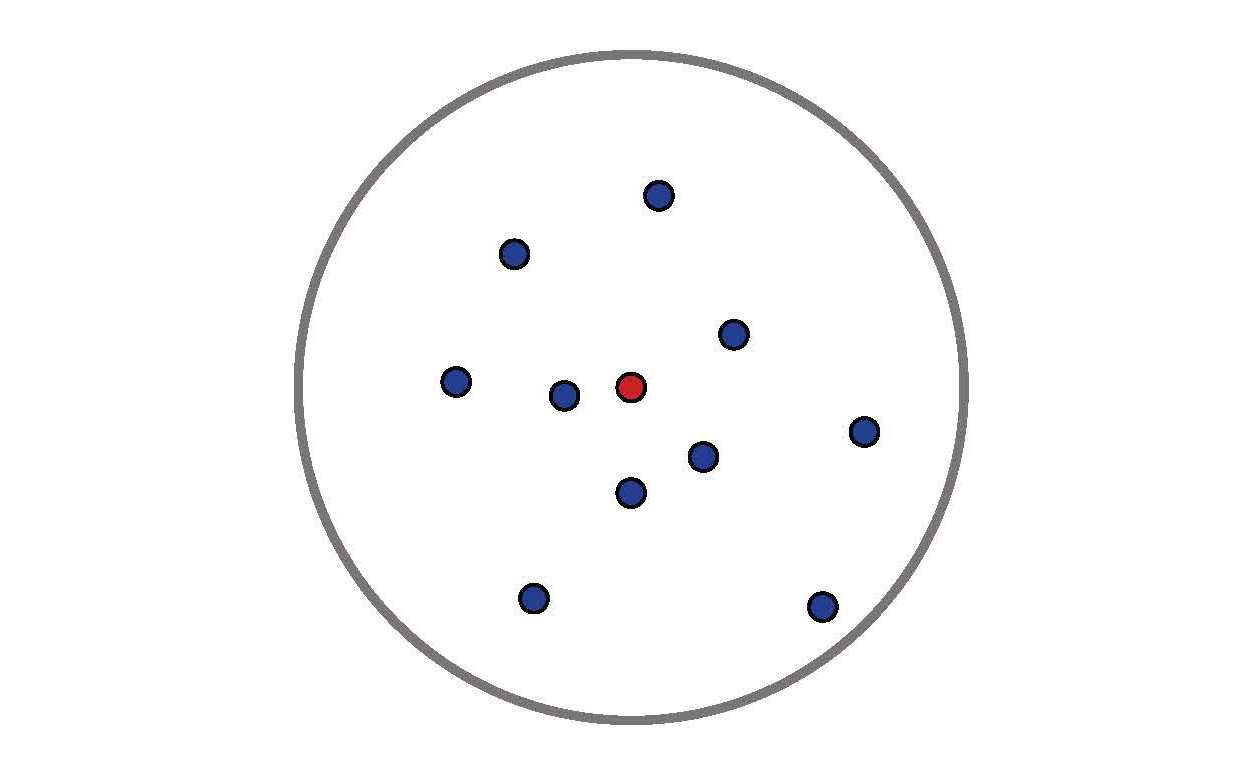}}%
\caption{Example situations of a normal movement and an undesired awakening. The event in question is coloured red.}
\label{fig:Movements}
\end{figure*} 

\subsection{Distance-based Algorithm}\label{sec:DistAlgo}
The aforementioned accuracy issues in IPS (noise type ACC), that particularly occur in narrow regions with bad signal coverage (e.g.~high-bay warehouses), should be considered for the construction of reliable distance-based algorithms. Therefore, we use an approach which relies on the density structure of past observations. The usage of the distance metric is the practicioners choice. We concentrate on the Euclidean distance which is between different events $i$ and $i'$ for a device $j$ noted by $\text{d}(\text{\textbf{u$_{i,j}$}}, \text{\textbf{u$_{i',j}$}})$ and defined as
\begin{equation*}
\sqrt{(x_{i,j} - x_{i',j})^2 +(y_{i,j} - y_{i',j})^2 + (z_{i,j} - z_{i',j})^2}.
\end{equation*}
For each event, we observe the position information of the past $k$ events and determine the maximal distance which exist to one of those. Expressed more mathematically, we apply a sliding count window \citep{ari2012data} over the whole array of all events of a device. A sliding count window buffers the past $k$ events and is successively sliding from the beginning to the end of the array. Consequently, there is an overlap by $k-1$ events between each step. In our context, we aim to classify the most recent event of each step (i.e. the event at the ``right" side of the current window). To do so, we apply the Euclidean distance between this event in question and any other event which lies in the window. Subsequently, we determine the maximum of those distances which we denote by mscw (maximum sliding count window distance). The whole procedure can be represented by 
\begin{equation*}
\text{mscw}_{i,j} = \max_{p \in \{1,..., k\}} \text{d}(\text{\textbf{u$_{i,j}$}}, \text{\textbf{u$_{i-p,j}$}}).
\end{equation*}
An event is then classified as AME if the criterion $\text{mscw}_{i,j} > r$ is fulfilled where $r$ is a pre-defined threshold. Note that no values of \text{mscw} are calculated for the first $k$ events. We consider those events as a burn-in phase of the process. Algorithm \ref{alg:dist} details the approach as a pseudo-code.

 \begin{algorithm}
 \caption{Distance Based Algorithm}
 \label{alg:dist}
 \begin{algorithmic}[1]
 \renewcommand{\algorithmicrequire}{\textbf{Input:}}
 \renewcommand{\algorithmicensure}{\textbf{Output:}}
 \REQUIRE Position events $i$ for a device $j$, radius $r$, Width of sliding count window $k$
 \ENSURE  Classification in AME or UAE
  \FOR{all $i = k+1, ..., n_j$}
  \STATE $\text{mscw}_{i,j} = \max_{p \in \{1,..., k\}} \text{d}(\text{\textbf{u$_{i,j}$}}, \text{\textbf{u$_{i-p,j}$}})$
  \IF{\normalfont $\text{mscw}_{i,j} > r$}
  \STATE $C_{i,j}$ = AME
  \ELSE 
  \STATE $C_{i,j}$ = UAE
  \ENDIF
  \ENDFOR
 \RETURN Classification = $\{C_{k+1,j}, ..., C_{n_j,j}\}$ 
 \end{algorithmic} 
 \end{algorithm}

Fig. \ref{fig:Movements} makes the criterion intuitively understandable. A sphere with a determined radius is drawn around the coordinates of each event. For an actual movement, we expect that only a few previous events lie within the applied sphere. Hence, an event is classified as AME if the amount of previous events, which lie within the sphere, is smaller than an applied threshold. Likewise, we follow that no meaningful distance was traveled and, hence, classify the signal as UAE, if this amount is at least as large as the threshold. Note that due to the usage of a sphere, it is possible to handle various system accuracies, large amount of noise, and different types of movements. 

\begin{figure}[!t]
	\centering
	\includegraphics[width = 3.2in]{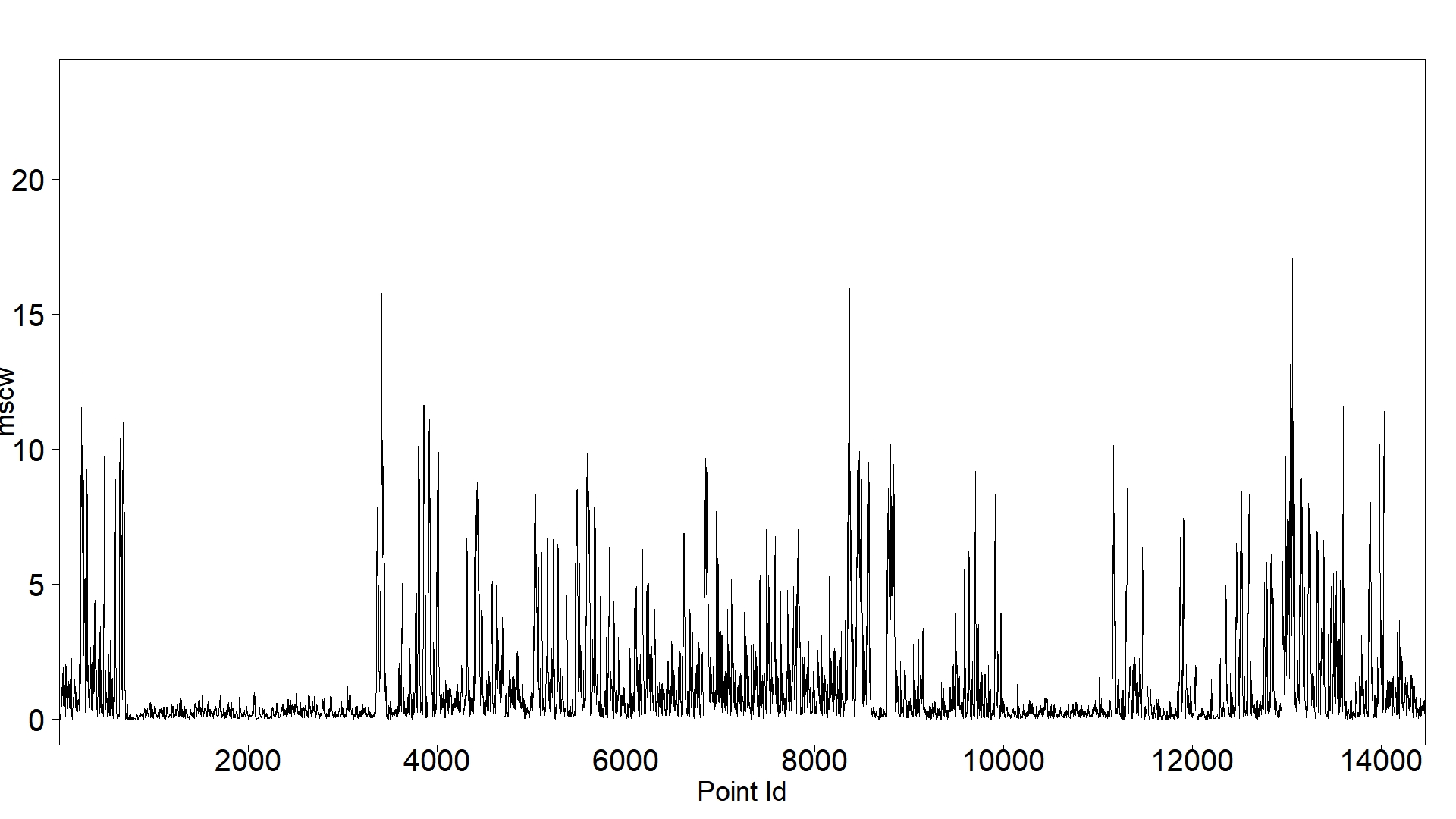}
\vspace{-0.5cm}
	\caption{mscw for each event of Device 199 for $k = 10$ (see also Section \ref{sec:Examp}).}
	\label{fig:Distance199}
\end{figure}

Obviously, the algorithm requires the choice of two input parameters: the radius of the sphere $r$ and a maximal number of signals $k$ that have to lie within the sphere such that the event in question is considered as AME. In this context, we note that the resulting classification is highly sensitive to the parameter $r$, since the applied radius has a large impact on the classification performance. Therefore, a careful parameter choice is crucial and it is obvious that an appropriate determination of $r$ depends on the distribution of the mscw over all events of a device. Thus, we propose to use graphical support and visualize those values in a bar plot for a fixed value of $k$ under the consideration of the order of the events which is given by their time of occurrence. See Fig. \ref{fig:Distance199} for an example.

Low values indicate UAE type events, whereas high values indicate AME type events. Under the consideration of the depicted structure, the user is now able to follow the traveled distances and to choose an appropriate bound which serves as the input parameter $r$. Every event with a value lower than this bound is then classified as UAE. In the given example, we observe two periods with values close to zero, while the values are way larger elsewhere. Hence, we can expect those events, that correspond to small values in Fig. \ref{fig:Distance199} to be of the UAE type. Consequently, we set $r = 1.5$ to obtain the desired classification in this example. Note that the general procedure of this approach is similar to the parameter tuning of the OPTICS algorithm \citep{ankerst1999optics}. However, we use a different criterion and do not require the update algorithm for the ordering of points as our order is naturally given by the time of occurrence. \\
In general, a distance-based approach is the most intuitive way to distinguish between actual movements and undesired awakenings. However, the resulting classification should not be overinterpreted. Slow movements of the AME type (i.e.~stops in classical stay point applications) might sometimes be deemed UAE, e.g.~storing processes or the formation of queues which naturally cause consecutive events with low traveled distances. It is therefore necessary to use other available information as well to support the distance-based classification in order to achieve a more reliable analysis.

\subsection{Time-based Algorithm}\label{sec:TimeAlgo}
Another perspective towards the problem is the consideration of the passed time between two events. Since no position coordinates are used and no distances are calculated, the advantage of this approach is the independence of systemic errors and measurement inaccuracies that depend on the quality of the area coverage.
The main assumption for the time-based criterion is that during actual movements the device sends the new position in small periods of time as the movement is quasi-continuous. Undesired awakenings, on the other hand, are assumed to be characterized by random or systematic occasional discrete awakenings of the device which can, e.g., be caused by large storage machines which are moving near to the device and shake it up such that it awakes.

We propose to use an algorithm which works as follows: For each event, the passed time between its occurrence and the occurrence of the event which was the $k$th-last observation is calculated. If this time difference is below a certain threshold, the point is labeled as AME. This is justified by the intention of smooth and fast movements in the production hall without larger breaks. If the threshold is exceeded, this hints to a rather time-discrete awakening structure and the signal is then classified as UAE. See Algorithm \ref{alg:Time} for a more formal definition. 

\begin{algorithm}[H]
\caption{Time Based Algorithm}
 \label{alg:Time}
\begin{algorithmic}[1]
\renewcommand{\algorithmicrequire}{\textbf{Input:}}
\renewcommand{\algorithmicensure}{\textbf{Output:}}
\REQUIRE Position events $i$ for a device $j$, Time Threshold $b$, Time Difference $k$
\ENSURE  Classification in AME or UAE
\FOR{all $i = k+1, ..., n_j$}
\STATE Diff = $t_{i,j} - t_{i-k,j}$
\IF{Diff $\leq b$}
\STATE $C_{i,j}$ = AME
\ELSE 
\STATE $C_{i,j}$ = UAE
\ENDIF
\ENDFOR
\RETURN Classification = $\{C_{k+1,j}, ..., C_{n_j,j}\}$ 
\end{algorithmic} 
\end{algorithm}

Similar to the distance-based algorithm, two input parameters have to be chosen. This includes a threshold $k$ which specifies between how many signals apart the time difference is calculated and a time threshold $b$ which defines the bound. We recommend a choice of parameters in a similar way like above, i.e. the parameter $k$ is directly determined, since it has little impact on the resulting classification. Furthermore, $b$ can then be chosen with the support of plots of the resulting time differences. Depending on the structure of the plot, $b$ can be set such that every event, for which the corresponding value lies below, is classified as AME and as UAE otherwise.

\begin{table*}[!t]
\caption{Characteristics of the aggregated classes.}
\centering
\renewcommand{\arraystretch}{1.2}
{\begin{tabular}{|l||c|c|l|}
\hline
Class & Distance-based & Time-based & Characteristics \\
\hline
\multirow{2}{*}{1} & \multirow{2}{*}{AME} & \multirow{2}{*}{AME} 
& movement detected by both criteria,\\  
& && long traveled distances with quasi-continuous signals\\ 
\hline
\multirow{3}{*}{2} &\multirow{3}{*}{AME}& \multirow{3}{*}{UAE}
& frequently time-discrete awakenings but with rather large\\
&&& distances between each other, hint towards undesired  \\
&&& awakenings in areas with large measurement errors\\
\hline
\multirow{2}{*}{3} & \multirow{2}{*}{UAE} & \multirow{2}{*}{AME}
&often quasi-continuous signals but small traveled distances,\\
&&& e.g. building of queues or running of big machines nearby \\
\hline
\multirow{2}{*}{4} & \multirow{2}{*}{UAE} & \multirow{2}{*}{UAE}
& no movement detected by both criteria, \\
&&& time-discrete awakenings near to each other \\
\hline
\end{tabular}}
\label{tab:char}
\end{table*} 

The advantage of time-based algorithms lies in their independence of distances which affects that measuring errors are ignored and natural causes for low traveled distances are acknowledged. As a result, signals of slow movements like in storage processes or queues can be recognized as actual movements. While being a meaningful support for distance-based algorithms, their sole usage, however, is not recommended either. The main disadvantage is the ignoring of long lasting disruptive factors which cause quasi-continuous awakenings of a device, e.g. the working of a production machine nearby for some period of time. The caused signals would be classified as AME although they are undesirably sent.

\subsection{Aggregation of the Algorithms}\label{sec:Agg}
As already mentioned, we deliberately propose to use the algorithms separately first and to combine their results afterwards. The separate analyzation enables us to generate a flexible interpretation scheme adapted to the described challenges and the combination guarantees to make use of the advantages and to overcome the disadvantages of both types of algorithms. Hence, all events are classified with the described algorithms first and the outcomes are then aggregated which result in four possible classes. Thus, it is ensured that both types of given information (i.e.~space and time) is used. Those classes and their individual characteristics are illustrated in Table \ref{tab:char}.
Obviously, Class 1 hints to actual movements, since the events are categorized as AME for the distance-based as well as for the time-based criterion. Contrary, Class 4 is a strong indicator for undesired awakenings. In terms of mixed-type classes, where one criterion classifies an event as AME and the other as UAE, the interpretation is more difficult and also dependent on the individual circumstances in the production hall. Class 2 often indicates undesired awakenings which occur with measurement inaccuracies larger than the set radius $r$ of the distance-based algorithm. A likely scenario for Class 3, on the other hand, are slow actual movements which are e.g. caused by storage processes or queues (i.e.~stops). If a big working machine is located near to the sensor, an undesired awakening due to its disruptive impact might also be a reason.  

\subsection{Practical Applications}
The algorithms are constructed based on unsupervised techniques which enables a user-friendly application in practice. Considering that the only assumption is the existence of the raw IPS data, this particularly involves an easy implementation, a rapid calculation and a directly available application. On a further note, the algorithms are constructed such that their usage is possible both in an offline and an online setup which means that they can deal with collected data of past events as well as with streaming data. This especially opens up the opportunity of a live monitoring in real-time IPS. 

\subsubsection{Offline Applications}
In offline applications, the proposed procedure is applied to the events of a historic database. This especially involves the classification of each single event into one of the classes described in Section \ref{sec:Agg} in order to get a comprehensive overview of the system behavior and capacity. Consequently, the user is able to validate effective processes working as planned which improves the transparency of the positioning system. Contrary, undesired awakenings can be detected and possible causes can be assigned with the help of the interpretations of the different classes given in Table \ref{tab:char}. Furthermore, it is a natural step to aggregate the signals to user-defined areas of the production hall. To do so, the user can classify each event of each device with the additional information of the located area. Subsequently, the distribution of the classes can be calculated for each area which gives an indication which regions are problematic in terms of system efficiency. We particularly recommend to analyze the data for a grid of reasonable size in order to generate a comprehensive map of the production hall. An example with use cases is illustrated in Section \ref{sec:Examp}.     

\begin{figure*}[!t]
\centering
\includegraphics[width = 5.1in, keepaspectratio]{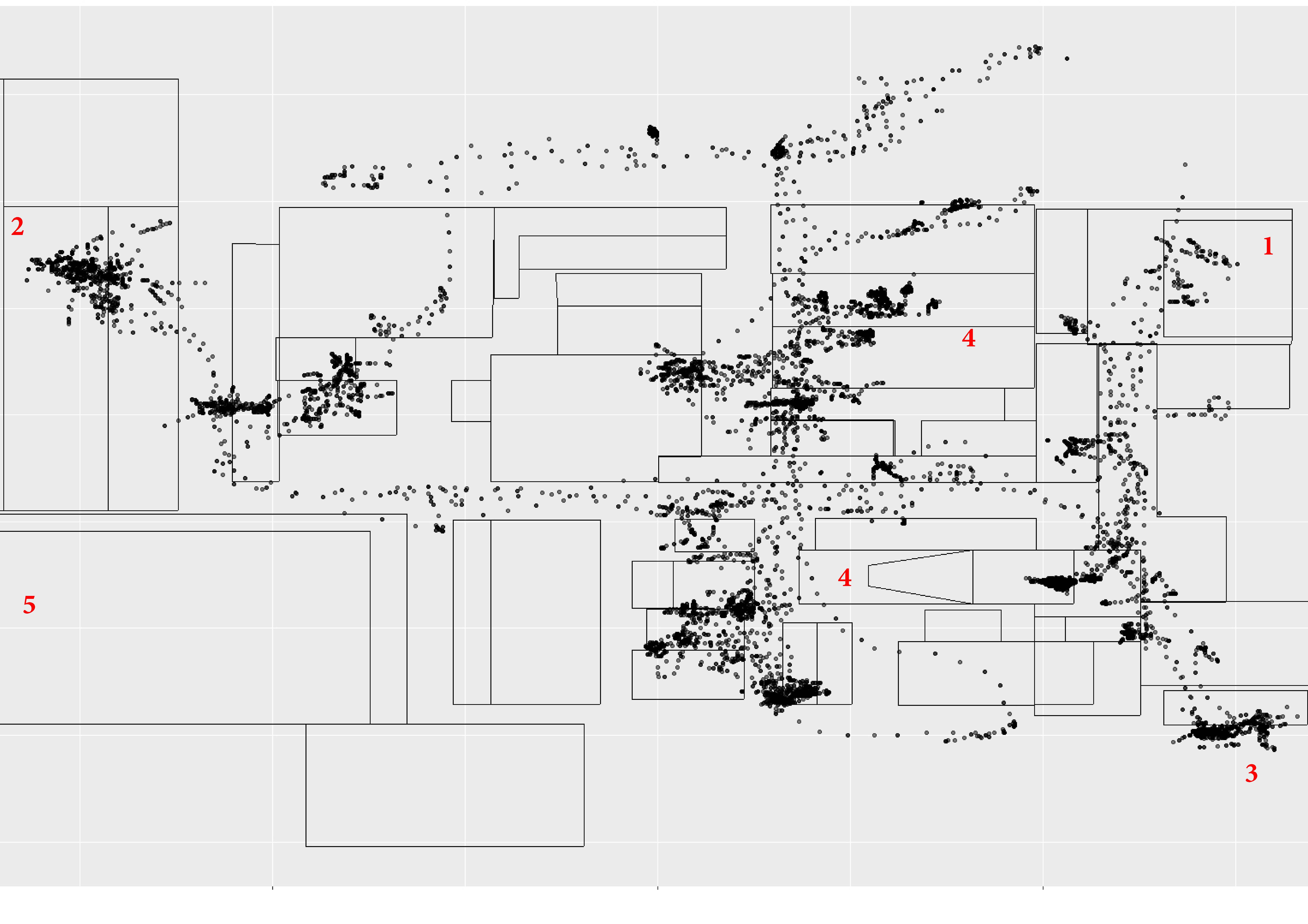}
\caption{All position events of device 199 in the production hall. The red numbers signalize main locations: 1 = big production machine, 2 = high-bay warehouse, 3 = device charging station, 4 = warehouses, 5 = assembly station}
\label{fig:GeoEnergy199}
\end{figure*}

\subsubsection{Online Applications}
Contrary, online applications deal with so-called streaming data. This means that each newly observed event is added to the already existing data set of past events and is, therefore, immediately analyzed and classified after its occurrence. This is why we also refer to this procedure as sequential analysis. It particularly enables the user to rapidly detect system errors or potential disrupting factors which have a negative impact on the efficiency. As a consequence, causes of interference can immediately be assigned and eliminated which reduces valuable time and cost resources. Similarly, an individual order can be monitored in real time to secure that important deadlines can be met (e.g. transport from warehouse to removal station) which increases the transparency of the production processes. If the according process is in danger to be delayed, this can be detected with the movement classification and a warning alarm can be triggered.

\section{Computational Experiments}\label{sec:Examp}
We now move on to the application of the proposed approaches in a real-world example to further illustrate the functioning and benefits of the proposed procedure. 

\subsection{Setup}
We use IPS data of our industry partner Arnold AG. The provided data comprise three-dimensional position coordinates of 401 devices. The considered period of time covers two months in which more than 3.5 millions of position events had been sent. Note that we worked completely unsupervised in this case study as the large volume of data makes the elevation of a ground truth hardly feasible. Thus, we evaluated our results with the help from experts who have daily experience with the processes and challenges of the IPS of our industry partner. Recall that our developed procedure aims to handle the described challenges of IPS data in production halls by offering a flexible interpretation scheme, which is, amongst others, possible due to the separate analyzation of both presented algorithms. The mentioned classical and more general stay-point algorithms in Section \ref{sec:relW} either use only distance-based algorithms or a joint analyzation of both types which results in a binary classification output. However, as already described, many circumstances (e.g.~noise type SLEEP, high dynamics in production halls) affect that a detected stop might not necessarily represent a stop in a classical sense and requires further investigation. 

 \begin{figure*}[!t]
	\centering
	\subfloat[Distance-based Algorithm]{\includegraphics[width = 5.1in, keepaspectratio]{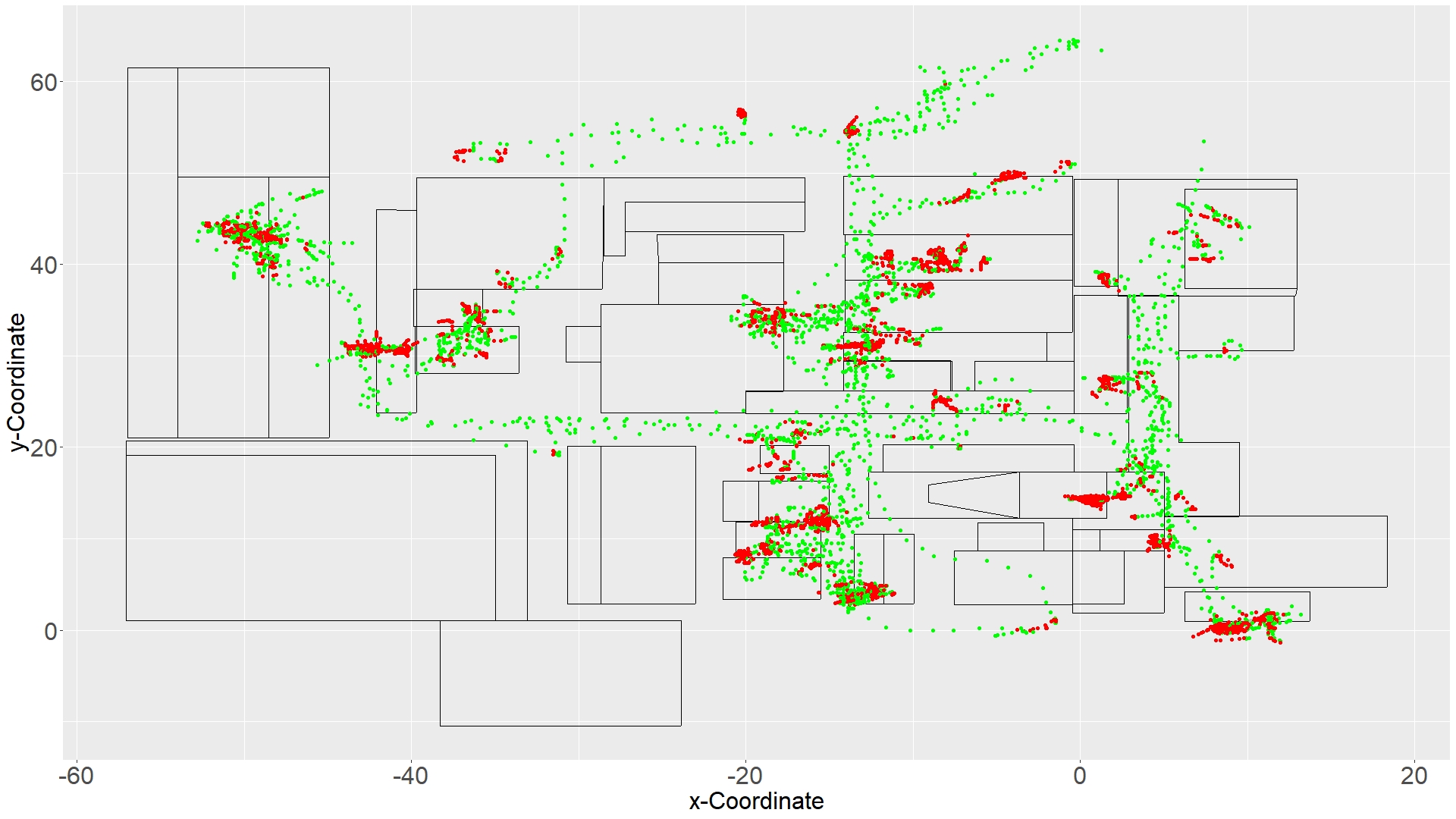}
\label{fig:ClassificationA}}%
\hfil
	\subfloat[Time-based Algorithm]{\includegraphics[width = 5.1in, keepaspectratio]{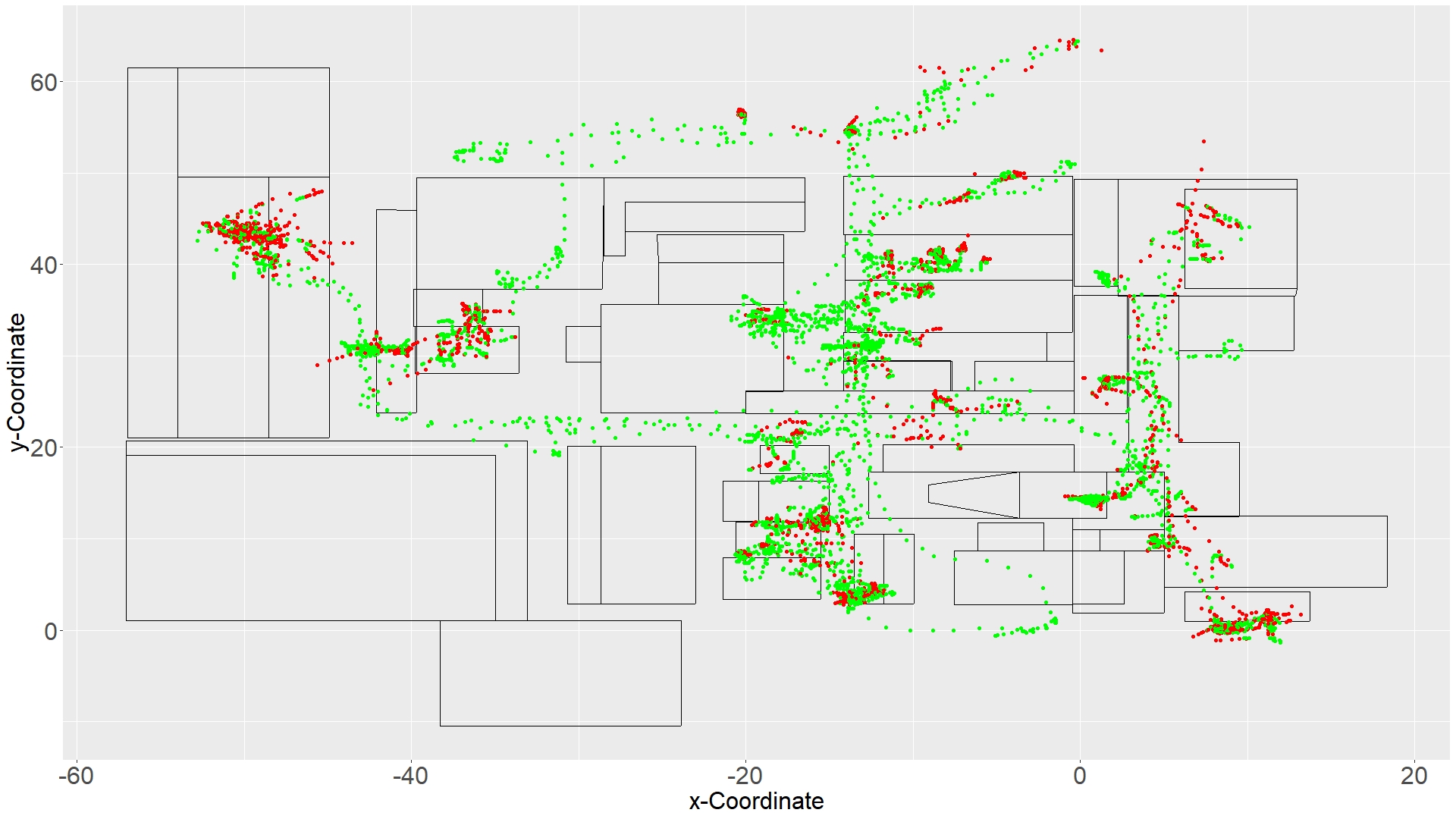}
\label{fig:ClassificationB}}%
	\caption{Classifications of the two algorithms with the applied parameter setups. Green dots signalize AME, red dots UAE.}
	\label{fig:Classifications}
\end{figure*}

\subsection{General Movement Structure}
In order to get a general impression of the system's behavior, we firstly apply both algorithms separately on the events for all devices. In the following, we give an example analysis of the particular device with ID number 199. In this context, Fig. \ref{fig:GeoEnergy199} gives an overview of the movement of the device. The layout of the production hall with the user-defined query regions is illustrated as well.

\begin{figure*}[!t]
	\centering
	\includegraphics[width = 5.1in, keepaspectratio]{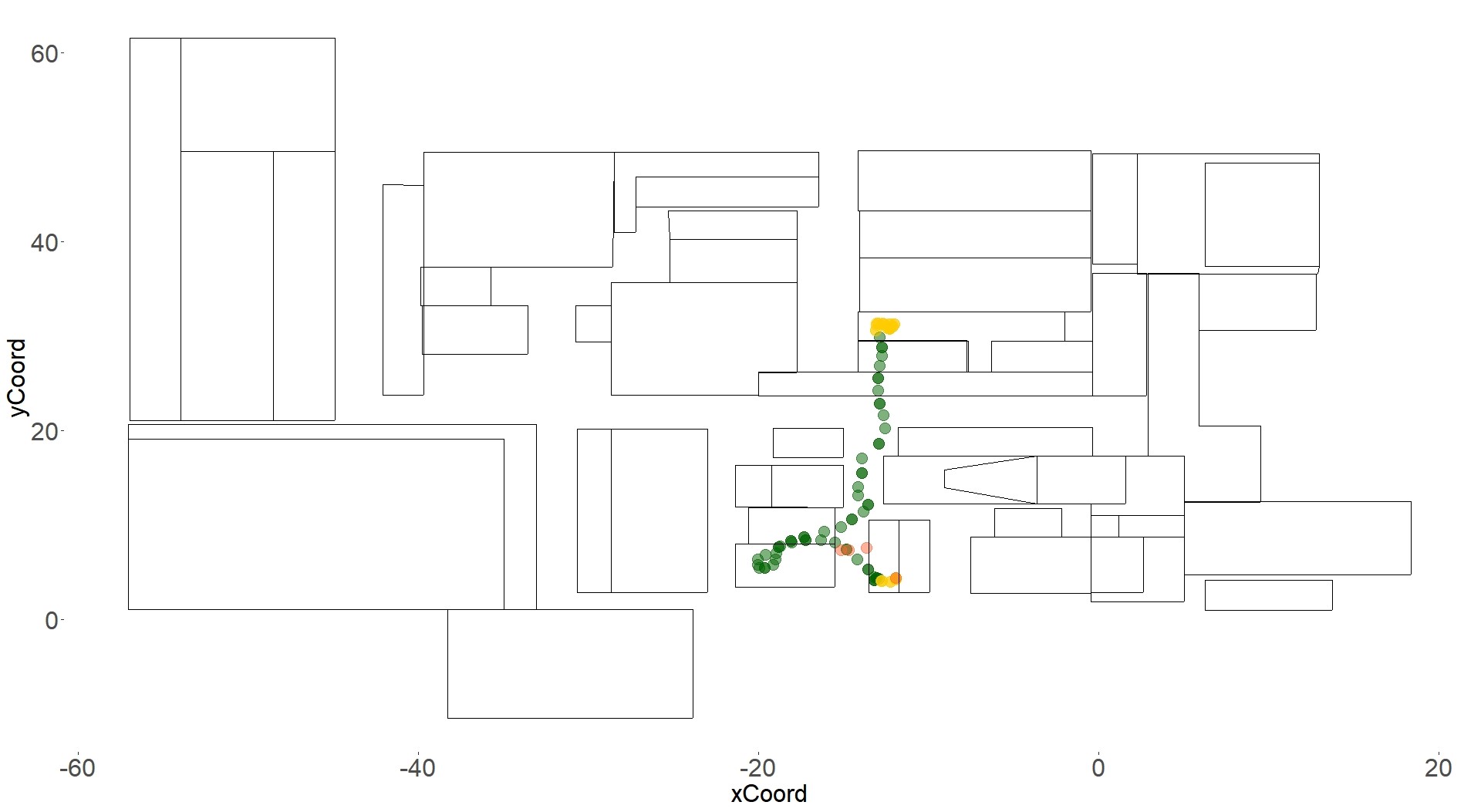}
	\caption{Monitoring of a transport procedure using the proposed algorithms. Colour Index: Green \crule[darkgreen]{0.3cm}{0.3cm} = Class 1, Orange \crule[orange]{0.3cm}{0.3cm} = Class 2, Yellow \crule[yellow]{0.3cm}{0.3cm} = Class 3, Red \crule[darkred]{0.3cm}{0.3cm} = Class 4. For the Definition of the classes see Table \ref{tab:char} in Section \ref{sec:Agg}.}
	\label{fig:OnlineApp}
\end{figure*}

Apart from many obvious movements along transport routes, there are also some regions of gatherings of points where short distances were traveled. This may hint to slower movements (i.e.~stops), which are typical for queues or storage processes in warehouses, but it might also be an indicator for undesirably sent data. To classify the events, we now apply the distance-based algorithm and set our parameters according to the already discussed visualization in Fig. \ref{fig:Distance199} in Section \ref{sec:DistAlgo} which shows the mscw distance for each event. We use $k = 10$ and $r = 1.5$ for our parameter setup which means that an event is deemed AME if the largest distance to one of its 10 predecessors exceeds 1.5 meters. Subsequently, we apply the time-based algorithm with $k = 10$ and $b = 15$ seconds. Hence, an event is classified as AME if its occurrence lies within 15 seconds of the 10th last event. This parameter setup seems reasonable considering the fact that a device sends position events every second if it awakes. However, a larger choice of $b$ might be justified as well if the user knows about a high fraction of actual movements which are disrupted by particular circumstances (e.g. due to queues).

The results of the distance-based and time-based classification criteria are both illustrated in Fig. \ref{fig:ClassificationA} and Fig. \ref{fig:ClassificationB}, respectively. 
At first glance, the resulting classifications look quite similar to each other. In particular, both algorithms seem to perform reasonably in terms of the detection of movements along the main transport routes. Differences between the classifications exist e.g.~for the region at the top right corner where the time-based algorithm detects UAE types more frequently. On the other hand, the distance-based algorithm is more conservative regarding the gathering of points at the middle part of the visualization. This behavior seems reasonable, since the signal coverage is expected to be better along the main parts of the production hall and lower in more narrow regions at the side parts. Hence, the noise is higher in the latter regions and probably succeeds the threshold of the distance-based algorithm more often which results in a classification dominated by AME types. 

\subsection{Application Example on Online Data}
We now move on to the application of our main procedure, i.e.~the combination of the results of both algorithms and the usage of the presented interpretation scheme. We consider online applications first by monitoring the already mentioned device No.~199. To do so, we use those events which are connected with an order to transport the device from a warehouse to a drilling station. We treat the according data sequentially to classify all incoming events immediately after their occurrence. We setup the monitoring procedure by triggering an alarm if either five consecutive events are classified into Class 4 or ten consecutive events into Class 2 or Class 3. Fig. \ref{fig:OnlineApp} summarizes the results. 

As we can see, the transport procedure was successful without any meaningful interruptions since nearly all events are classified into Class 1. Hence, the order was never in danger to fail and the user was able to follow and validate the successful execution at all time. The monitoring only signaled an alarm at the very beginning of the process due to some consecutive events belonging to Class 3. However, this is characteristic for storage processes as explained in Section \ref{sec:Agg}. If the user wishes to be not informed about such alarms (i.e~which are likely sent due to natural causes), the threshold for consecutive events of Class 2 or Class 3 can be set more liberal.

\begin{figure*}[!t]
	\centering
	\includegraphics[width = 5.5in, keepaspectratio]{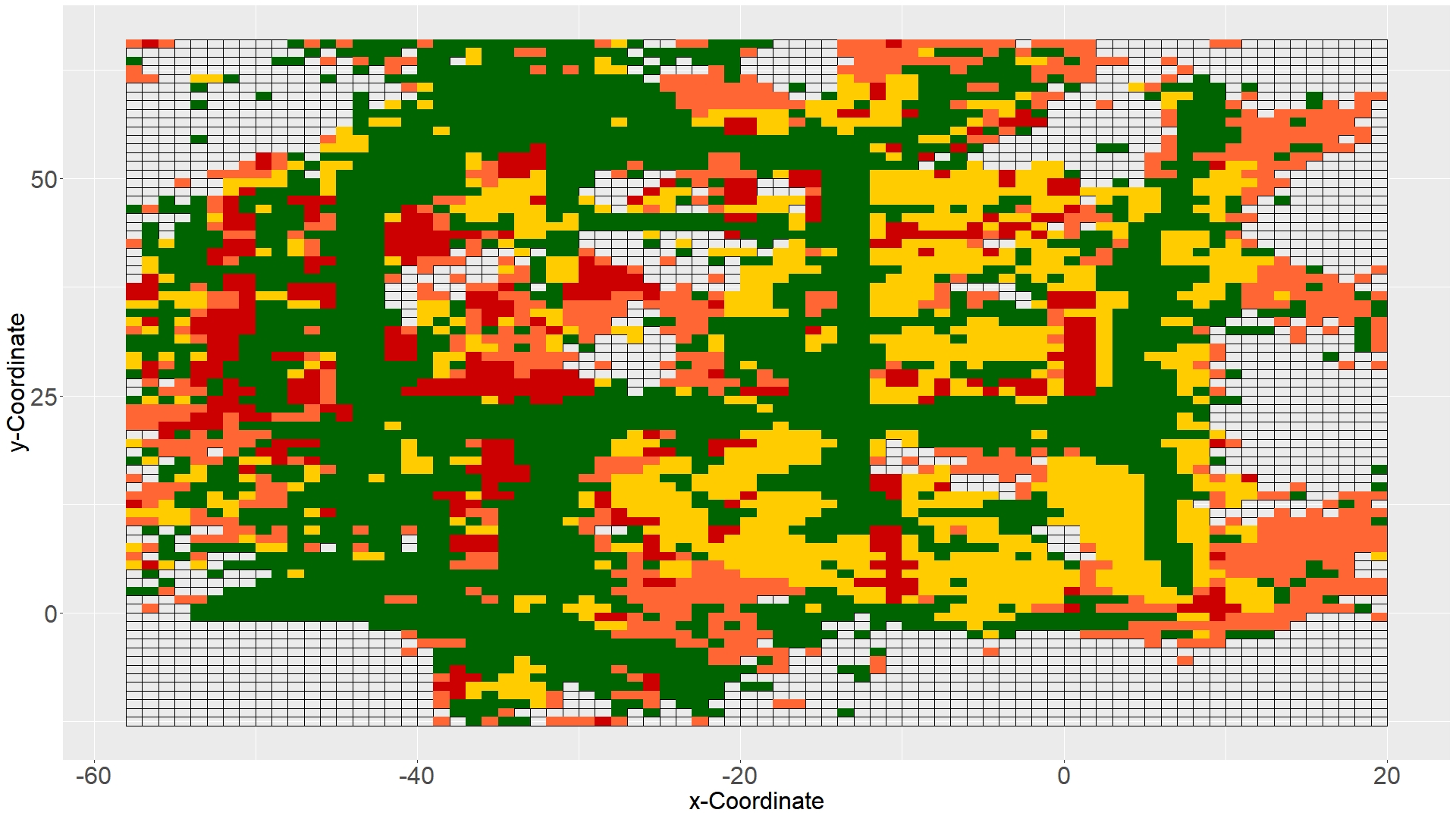}
	\caption{The most dominant of the 4 aggregated classes for each grid cell. Colour Index: Green \crule[darkgreen]{0.3cm}{0.3cm} = Class 1, Orange \crule[orange]{0.3cm}{0.3cm} = Class 2, Yellow \crule[yellow]{0.3cm}{0.3cm} = Class 3, Red \crule[darkred]{0.3cm}{0.3cm} = Class 4. For the Definition of the classes see Table \ref{tab:char} in Section \ref{sec:Agg}.}
	\label{fig:AreaCl}
\end{figure*}

\subsection{Application Example on Offline Data}
Regarding offline applications, we repeat the application of the procedure for each device. In the end, each event in the observed period of time is classified into one of the four classes described in Section \ref{sec:Agg}. To achieve an overall system analysis, we further aggregate the results to spatial production areas to get a comprehensive overview of the system's behavior in different areas of the production hall. To do so, we span a grid across the layout of the production hall such that each cell covers a space of one square meter. In the end, we analyze each grid cell by assigning the class which occurs most frequently among all events that are located in its covered area. Fig. \ref{fig:AreaCl} shows the resulting map where each Class is characterized by its own colour. A grid cell is coloured gray if less than five events took place in the corresponding area.

The map gives a comprehensive overview over the production hall and is able to detect its main characteristics. This can be compared and justified with the depicted layout in Fig. \ref{fig:GeoEnergy199}. The green coloured cells signalize the main transport routes where the majority of events are classified as AME for both types of algorithms. Contrary, the red cells are characteristic for a dominance of UAE types for both algorithms. In particular, this takes place in regions which are located directly next to the busy transport routes where stored devices frequently awake due to passing vehicles. Furthermore, the battery charging station of the devices is coloured red as well, since there naturally occur many systemic errors. Regarding the mixed-types, we observed rather high correlations of warehouses with the yellow marked cells that stand for a distance-based UAE and a time-based AME classification. This characterizes the slower but time-consecutive movements of storage processes in a warehouse (i.e.~stops or stay-points). The orange cells indicate a domination of distance-based AME and time-based UAE classification. This especially occur in regions located at the side parts of the hall where coverage is expected to be worse and noise influences the quality of the distance-based algorithm which gets too liberal. An example are the narrow parts of the high-bay warehouse where stored devices often undesirably awake, because it is quite busy there. However, the system accuracy suffers in this region and therefore the distance-based criterion often signalizes AME. This problem can e.g.~be solved by adapting the sensibility threshold of the sensors in those regions. 
Thus, the map not only helps for a visualization of the actual movement structure in the production hall, but also for the detection of problematic system regions which are characterized by undesired awakenings (particularly red and orange cells). This enables the user to investigate and eliminate possible causes in order to achieve gains in energy consumption as well as in system efficiency. Furthermore, the map may serve as an update or validation process of rather old layout maps and is automatically able to visualize possible layout changes.

\section{Conclusion} 
While IPS are developed further and produce more valuable but complex data, their analysis gets more challenging. We proposed an approach to reliably detect movement and to analyze the movement structure in the production hall. To do so, we aimed to distinguish between events, which represent actual movements and events that were undesirably sent due to disrupting factors. Our proposed procedure is directly applicable, enables the handling of various system accuracies and intensities of noise and is particularly designed to handle the challenges of IPS in production halls including noise type SLEEP. Since we use a distance-based as well as a time-based criterion separately, we achieve a flexible classification which can be further applied to user-defined query regions in the production hall. This not only helps for a more efficient usage of the positioning system, but also for a better understanding of the own processes and the detection of anomalies in particular regions of the production hall. Furthermore, we proposed to summarize the results with visual analytics in a grid structure which enables us to generate a layout map of the production hall with the sole usage of the raw IPS data. For future researches, our results can e.g.~be used for continuous updates of the map of the underlying production hall or for the preparation of a semi-supervised technique to comprehensively analyze the causes of a rather bad performance in problematic areas.  

\section*{Acknowledgements}
This work was supported by the Mercator Research Center Ruhr (MERCUR) under Grant PR-2019-0019

\section*{Disclosure Statement}
The authors report there are no competing interests to declare.

\bibliographystyle{asmems4}
\bibliography{References}

\end{document}